\documentclass[10pt, a4paper]{article}
\usepackage{lrec2022} 
\usepackage{multibib}
\newcites{languageresource}{Language Resources}
\usepackage{graphicx}
\usepackage{amsmath,amssymb,amsfonts}
\usepackage{tabularx}
\usepackage{multirow}
\usepackage{soul}
\usepackage{caption}
\usepackage{subcaption}
\usepackage{titlesec}
\titleformat{\section}{\normalfont\large\bfseries\center}{\thesection.}{1em}{}
\titleformat{\subsection}{\normalfont\SmallTitleFont\bfseries\raggedright}{\thesubsection.}{1em}{}
\titleformat{\subsubsection}{\normalfont\normalsize\bfseries\raggedright}{\thesubsubsection.}{1em}{}
\renewcommand\thesection{\arabic{section}}
\renewcommand\thesubsection{\thesection.\arabic{subsection}}
\renewcommand\thesubsubsection{\thesubsection.\arabic{subsubsection}}

\usepackage{epstopdf}
\usepackage[utf8]{inputenc}

\usepackage{hyperref}
\usepackage{xstring}

\usepackage{color}

\usepackage{multirow}

\title{CinPatent: Datasets for Patent Classification}

\name{Minh-Tien Nguyen$^{1,2}$, Nhung Bui$^{1}$, Manh Tran-Tien$^{2}$, Linh Le$^{3}$, Huy-The Vu$^{2}$} 

\address{$^{1}$Cinnamon AI, 10th floor, Geleximco building, 36 Hoang Cau, Dong Da, Hanoi, Vietnam. \\
        \{ryan.nguyen, joanna\}@cinnamon.is\\
         $^{2}$Hung Yen University of Technology and Education, Hung Yen, Vietnam. \\
         \{tiennm, manhtt, thevh\}@utehy.edu.vn\\
         $^{3}$The University of Queensland, Australia.\\
         linh.le@uq.edu.au\\}

\abstract{
Patent classification is the task that assigns each input patent into several codes (classes). Due to its high demand, several datasets and methods have been introduced. However, the lack of both systematic performance comparison of baselines and access to some datasets creates a gap for the task. To fill the gap, we introduce two new datasets in English and Japanese collected by using CPC codes. The English dataset includes 45,131 patent documents with 425 labels and the Japanese dataset contains 54,657 documents with 523 labels. To facilitate the next studies, we compare the performance of strong multi-label text classification methods on the two datasets. Experimental results show that AttentionXML is consistently better than other strong baselines. The ablation study is also conducted in two aspects: the contribution of different parts (title, abstract, description, and claims) of a patent and the behavior of baselines in terms of performance with different training data segmentation. We release the two new datasets with the code of the baselines.
 \\ \newline . }

\begin{document}

\maketitleabstract

\section{Introduction}

Patent documents are an immense knowledge source for both research and innovation communities. The rapidly increasing number of patent documents poses enormous challenges for organizing and mining this source in an effective manner \cite{Krestel-Patent-analysis-WPI-21}. In practice, when arriving at the office, a new patent application is firstly categorized by using classification codes \cite{Tran-CPC-Codes-ICMIKE-17,Tang-MLP-GCN-AAAI-20,Lee-UPTO-3M-20,Pujari-Mutil-task-Transformers-ECIR-21}. Since this task is manually done by examiners, it requires domain expertise, is time-consuming, and labor-expensive.

To reduce the paper work of domain experts, patent documents can be automatically assigned into pre-defined codes by using classifiers \cite{Lim-IPC-ICADMA-16,Tran-CPC-Codes-ICMIKE-17,Tang-MLP-GCN-AAAI-20,Lee-UPTO-3M-20,Pujari-Mutil-task-Transformers-ECIR-21}. Patent classification can be seen as multi-label document classification \cite{Tang-MLP-GCN-AAAI-20,Lee-UPTO-3M-20}. Given a patent document, classification models need to assign the patent into one or more codes (classes). To build the classifiers, several methods have been used, such as feature engineering \cite{Tran-CPC-Codes-ICMIKE-17,Hepburn-ALTA-18}, deep neural networks \cite{fastxml,Li-DeepPatent-Scientometrics-18,You2019AttentionXML:Classification,Tang-MLP-GCN-AAAI-20}, or pre-trained models \cite{Lee-UPTO-3M-20,Pujari-Mutil-task-Transformers-ECIR-21}. To leverage the development of patent analysis, several patent datasets have been also created. Table \ref{tab:observation} shows the observation of relevant datasets for patent classification. We can observe that the datasets are different in terms of the number of training samples and codes. In the past, most of patent datasets were derived from CLEF-IP \cite{Li-DeepPatent-Scientometrics-18}. Recently, UPTO-3M \cite{Lee-UPTO-3M-20} seems to be the largest dataset for patent classification. The dataset includes around 3M patent documents collected from Google Patents Public Datasets\footnote{https://console.cloud.google.com/marketplace/product/\\google\_patents\_public\_datasets/google-patents-public-data?project=api-project-502040931957} based on Big Query statements.\footnote{https://console.cloud.google.com/bigquery?p=patents-public-data} Other patent datasets were also created for other languages such as KPatent \cite{Lim-IPC-ICADMA-16} for Korean and FIPS \cite{Yadrintsev-patent-search-JPC-18} for Russia. (ALTA\footnote{http://www.alta.asn.au/events/sharedtask2018/description.html} and WIPO-alpha\footnote{https://www.wipo.int/classifications/ipc/en/ITsupport/\\Categorization/dataset/wipo-alpha-readme.html}).

\begin{table}[h!]
\caption{The observation of relevant datasets for patent classification.}\label{tab:observation}
\resizebox{\columnwidth}{!}{%
\begin{tabular}{lccccc}
\hline
\textbf{Dataset} & \textbf{Train} & \textbf{Test} & \textbf{Codes} & \textbf{Lang} \\ \hline
ALTA & 3971 & 1000 & 8 & EN \\
WIPO-alpha & 46,324 & 28,926 & 451 & EN \\
KPatent \cite{Lim-IPC-ICADMA-16} & --- &  564,793 & 630 & KR \\
CPC \cite{Tran-CPC-Codes-ICMIKE-17} & 305,895 &  131,97 & 613 & EN \\
MPatent \cite{Hu-MPatent-Sustainability-18} & 72,532 & 2679 & 96 & EN \\
FIPS \cite{Yadrintsev-patent-search-JPC-18} & 543,000 &  232,800 & 650 & RU \\
CLEF-IP \cite{Li-DeepPatent-Scientometrics-18} & 742,097 &  1350 & 622 & EN \\
UPTO-2M \cite{Li-DeepPatent-Scientometrics-18} & 1,950,247 & 49,900 & 637 & EN \\
UPTO-3M \cite{Lee-UPTO-3M-20} & 3,050,615 & 49,670 & 656 & EN \\
CIRCA \cite{Tang-MLP-GCN-AAAI-20} & 36,420 & 9106 & 555 & EN \\
 \hline
\end{tabular}%
}


\end{table}

Although patent classification rapidly grows with a lot of attention for both methods and datasets \cite{fastxml,Li-DeepPatent-Scientometrics-18,You2019AttentionXML:Classification,Tang-MLP-GCN-AAAI-20,Pujari-Mutil-task-Transformers-ECIR-21} and datasets \cite{Li-DeepPatent-Scientometrics-18,Lee-UPTO-3M-20,Tang-MLP-GCN-AAAI-20}, we argue that there still exists the gap in the growth. For the methods, evaluation is segmented in which classifiers are separately evaluated on different datasets. It leads to the lack of systematic comparison among strong baselines. For the datasets, they are sometimes hard to access, not well centralized, and some datasets are not available, e.g. CIRCA \cite{Tang-MLP-GCN-AAAI-20}. To fill the gap, we introduce two new patent datasets for English and Japanese. Patent documents are collected from the Google Patent database. The English dataset is to test classification methods in a common setting while the Japanese dataset is created to deal with low-resource languages. Based on the creation, we make a systematic comparison of strong multi-label classification methods. We hope that the new datasets and systematic comparison will leverage the task of patent classification.
The contributions of our paper are:
\begin{itemize}
    \item We collect and release two new patent datasets for evaluating patent classification models.\footnote{Data and code: https://github.com/Cinnamon/CinPatent} The first dataset, named CinPatent-EN, contains 45,131 English patents tagged with 425 classes, while the other, named CinPatent-JA, has 54,657 Japanese patents assigned with 523 classes.
    
    \item We experiment with both datasets by using eight strong multi-label text classifiers which are tree-based and deep neural network-based models. Evaluation results show that AttentionXML outperforms the other methods with large margins.
    
    \item We investigate which parts (title, abstract, description, and claims) of patents are more informative to the classifiers. Results indicate that using title+abstract+description+claim1 outputs better results. Training with different data segmentation also shows the behavior of the baselines.
\end{itemize}

\section{Related Work}
\paragraph{Data}
As shown in Table \ref{tab:observation}, UPTO-3M \cite{Lee-UPTO-3M-20} and CIRCA \cite{Tang-MLP-GCN-AAAI-20} are two recent datasets for patent classification. For UPTO-3M \cite{Lee-UPTO-3M-20}, patent documents were collected from Google Patents Public Datasets by using BigQuery with 656 classes. The authors used both IPC and CPC codes for the collection. The highest F-score was 66.83\% by using CPC+Claim when tested with PatentBERT \cite{Lee-UPTO-3M-20} on 49,670 documents. For CIRCA \cite{Tang-MLP-GCN-AAAI-20}, 45,526 patent documents were collected by using IPC codes with 555 classes. The classification model using a graph structure achieved 78.30\% of $P@1$. Two other big datasets are UPTO-2M \cite{Li-DeepPatent-Scientometrics-18} and CLEF-IP \cite{Li-DeepPatent-Scientometrics-18}. The UPTO-2M includes nearly 2M patent documents for training and 49,900 for testing. The best model on this dataset is DeepPatent which uses CNN \cite{Li-DeepPatent-Scientometrics-18}. It achieved 73.88\% of precision with no $F1@ 1$ disclosed. The CLEF-IP\footnote{http://ifs.tuwien.ac.at/~clef-ip/} was mostly used in the past for patent classification. The dataset contains patents between 1978 and 2009 with IPC codes. The best method is DeepPatent \cite{Li-DeepPatent-Scientometrics-18} with 83.98\% of precision. Other datasets in other languages can be also found such as KPatent \cite{Lim-IPC-ICADMA-16} for Korean and FIPS \cite{Yadrintsev-patent-search-JPC-18} for Russia. We share the trend of patent classification by introducing two new datasets for English and Japanese.

\paragraph{Methods}
There is also a growing number of methods for patent classification. The methods range from traditional methods by using feature engineering \cite{Lim-IPC-ICADMA-16,Tran-CPC-Codes-ICMIKE-17,Hepburn-ALTA-18,Yadrintsev-patent-search-JPC-18} or deep neural networks \cite{Li-DeepPatent-Scientometrics-18,Hu-MPatent-Sustainability-18}. Experimental results indicated that these methods achieved promising results. There are several methods of multi-label document classification which can be also adapted to patent classification \cite{fastxml,Jain2016ExtremeApplications,Prabhu2018Parabel:Advertising,Liu2017DeepClassification,Xiao2019Label-SpecificClassification,You2019AttentionXML:Classification}. The recent success of pre-trained models also leverages patent classification \cite{Lee-UPTO-3M-20,Pujari-Mutil-task-Transformers-ECIR-21}. For example, PatentBERT \cite{Lee-UPTO-3M-20} utilizes BERT for fine-tuning the model on the UPTO-3M dataset. By using patent claims, PatentBERT achieves the best result of 66.83\% in terms of F-score. We extend the idea of using pre-trained models by comparing PatentBERT with other strong baselines, e.g. AttentionXML proposed for multi-label document classification.

\section{The New Datasets}

\subsection{Data Collection}
\paragraph{CPC structure}
There are two common code systems: CPC (Cooperative Patent Classification) and IPC (International Patent Classification) for patent management. For data collection, we collected data assigned with CPC labels since the CPC system is more specific and detailed than the IPC system and a growing number of national patent offices has also followed the CPC system \cite{Degroote-Analysis-WPI-18}.

CPC codes are hierarchically structured to define the finer scope of a patent. For example, the code ``A01B33/00'' denotes a patent in section A (``human necessities''), class 01 (``agriculture; forestry; animal husbandry; hunting; trapping; fishing''), subclass B (``soil working in agriculture or forestry; parts, accessories of agricultural machines or implements, in general''), etc. Our datasets contain labels up to the subclass level, i.e., the first four characters of a patent code.

\paragraph{Patent structure}
A patent typically contains five parts: title, abstract, claims, description, and drawings. The innovation is summarized in the patent abstract. Patent claims determine the scope of protection for the invention. Patent description or specification provides a detailed explanation of the invention, which is visually enhanced by multiple drawings. Our datasets only include four of them, excluding the drawings.

\paragraph{Collection}
Patent documents were collected from Google Patent. For English, we directly obtained a subset from Google Patent Public Datasets.\footnote{https://github.com/google/patents-public-data} For Japanese, we scraped them from the Google Patent website.\footnote{https://www.google.com/patents}

\subsection{Data Preprocessing}
Besides title, abstract, claims, and description, we introduce the first claims extracted from patent claims to our datasets. In patent drafting, the first claims are required to set out the distinctive features of the invention, which to be the most informative among all claims. For the first claim extraction, we took the first paragraph in patent claims that begins with ``1''.

\paragraph{Tokenization} English and Japanese have different textual units: the former uses words whereas the later uses characters. This difference requires different methods for tokenization. In our datasets, we normalized the unit separation with a single blank space. Particularly, we first obtained a list of units with NLTK for English text and Fugashi \cite{Mccann2020FugashiPython} for Japanese text. The units were concatenated by the specified separator to create universal tokenized datasets.

\begin{figure}[h!]
    \centering
    \includegraphics[width=\columnwidth]{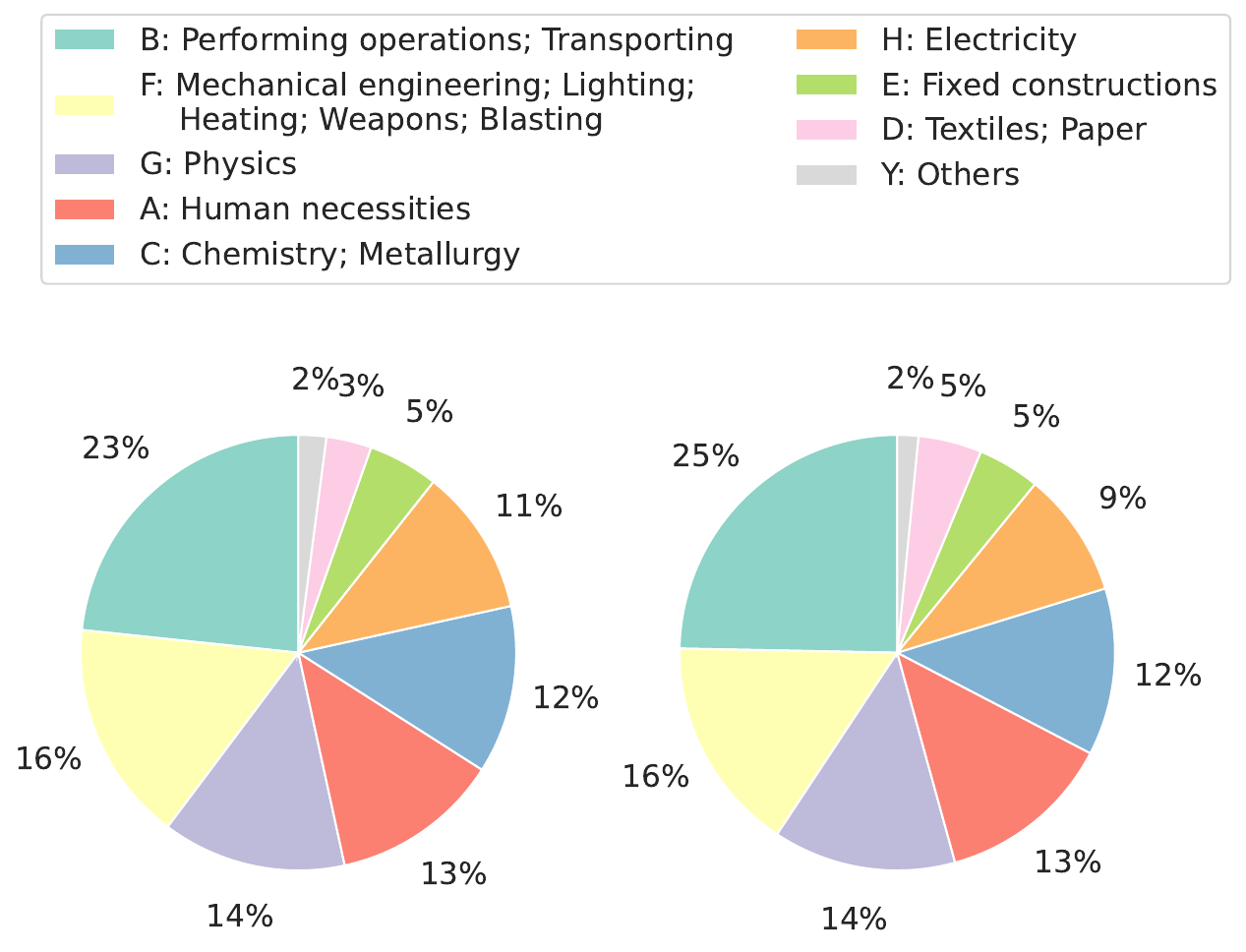}
    \caption{Label distribution at the section level in CinPatent-EN (left) and CinPatent-JA (right).}
    \label{fig:label_distribution}
\end{figure}

\paragraph{Class imbalance} Patent classes are imbalanced at multiple levels. Figure \ref{fig:label_distribution} illustrates label distribution in our datasets at the section level, i.e., the first character in the patent code. On the other hand, multi-label patents pose unique challenges for data balancing. In our datasets, we adopted a greedy undersampling technique with a balance threshold of 200 samples per label. Specifically, we first sorted the codes by their number of related samples. Starting from the codes with the least samples, we inherently selected all samples that are previously chosen in visited codes. The remaining samples were randomly selected until the threshold was reached. Intuitively, our approach gives higher priority to less frequent classes during undersampling.

One notable problem for our approach happens with popular codes, i.e., the codes that co-occur with many other codes, creating high peaks in the balanced label distribution. To mitigate this problem, we iteratively discarded the most frequent codes after a round of balance and manually observe the label distribution.

\subsection{Data Characteristics}
\begin{figure}[h!]
    \centering
    \centering
    \begin{subfigure}[b]{0.48\columnwidth}
        \includegraphics[width=\columnwidth]{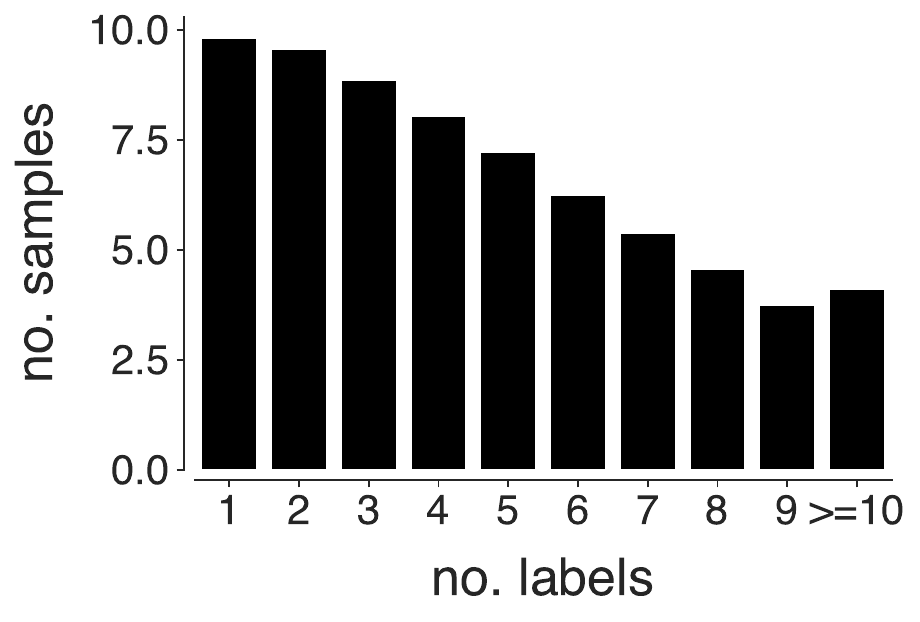}
    \end{subfigure}
    \begin{subfigure}[b]{0.48\columnwidth}
        \includegraphics[width=\columnwidth]{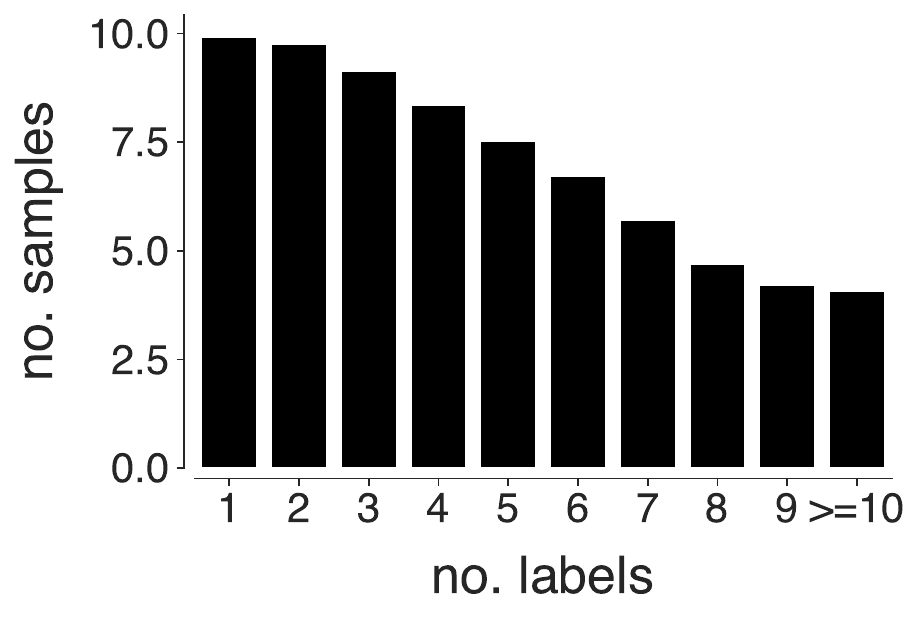}
    \end{subfigure}
    \caption{Number of labels per sample in CinPatent-EN (left) and CinPatent-JA (right).}
    \label{fig:labels_per_sample}
\end{figure}

We release our datasets for English, denoted as CinPatent-EN, and Japanese, denoted as CinPatent-JA. CinPatent-EN contains over 45k patents assigned in 425 CPC codes and CinPatent-JA contains over 54k patents assigned in 523 CPC codes. On average, a patent has 1-3 labels and the number of samples for labels is balanced around 200. We depict the number of labels per sample in Figure \ref{fig:labels_per_sample}. Patent descriptions provide the most textual information with more than 6k units per English sample and 9k per Japanese sample. Despite their length, we might employ its inverted pyramid structure for efficient document embedding. We provide overall statistics of our datasets in Table \ref{tb:data_stat} and statistics by different parts in Table \ref{tb:data_stat_by_part}.

\begin{table}[h!]
    \centering
    \caption{Data statistics of CinPatent-EN and CinPatent-JA. $N$ denotes number of samples, $D$ denotes the total number of tokens in a sample, $L$ denotes the total number of labels, ${\bar{L}}$ denotes the average number of labels per sample, ${\hat{L}}$ denotes the average number of samples per label.}\label{tab:data}
    \label{tb:data_stat}
    \resizebox{1\columnwidth}{!}{
    \begin{tabular}{l c c c c c c}
    \hline
     \multicolumn{2}{c}{\textbf{Datasets}}    &$N$ & $D$ &$L$ & $\bar{L}$ & $\hat{L}$  \\ \hline
     \multirow{3}{*}{CinPatent-EN} &Train & 35,663 & 7,520 & 425 & 2.09 & 175.76  \\
                                &Dev & 4,825 & 7,498 & 425 & 2.04  & 23.11  \\
                                &Test & 4,643 & 7,495 & 425 & 2.09 & 22.82 \\ \hline
                                
     \multirow{3}{*}{CinPatent-JA} &Train & 43,700 & 10,151 & 523 & 2.16  & 180.44  \\
                                &Dev & 5,504 & 11,160 & 523 & 2.18  & 22.98  \\
                                &Test & 5,453 & 11,797 & 523 & 2.26  & 23.52 \\ \hline

    \end{tabular}}
\end{table}

In Table \ref{tab:data}, Japanese patents are longer than English patents. The average number of labels per sample is quite similar, around two labels per sample. The average number of samples per label is also similar, showing the same characteristics of the two datasets.


\paragraph{Missing fields}
We observe 94.34\% Japanese patents do not have abstract and 7.03\% missing first claims. The missing data urges for the combination of textual fields in patent classification. We provide complete statistics of missing fields in Table \ref{tb:data_stat_by_part}.

\begin{table}[h!]
    \centering
    \caption{Data statistics by different parts in CinPatent-EN and CinPatent-JA: (\textbf{1}) title, (\textbf{2}) abstract, (\textbf{3}) claim1, (\textbf{4}) all claims, and (\textbf{5}) description. We use ``--" to denote parts with no missing values.}
    \label{tb:data_stat_by_part}
    \resizebox{1\columnwidth}{!}{
    \begin{tabular}{l c c c c c c}
    \hline
     \multicolumn{2}{c}{\textbf{Datasets}}   & \textbf{1} & \textbf{2} & \textbf{3} & \textbf{4} & \textbf{5}   \\ \hline
     \multirow{2}{*}{CinPatent-EN} &No. tokens & 7.29 & 131.05 & 154.68 & 955.88 & 6420.69  \\
                                &Missing & -- & 0.69\% & 0.16\% & -- & --  \\ \hline
                                
     \multirow{2}{*}{CinPatent-JA} &No. tokens & 17.6 & 10.97 & 137.54 & 1026.45 & 9361.97  \\
                                &Missing & -- & 94.34\% & 7.03\% & 0.27\% & 0.27\%  \\ \hline

    \end{tabular}}
\end{table}



\paragraph{Data splitting} 
We created training, development, and testing sets with 80:10:10 ratio. Considering multi-label samples, we opt to use iterative stratification in Scikit-multilearn to ensure stratified sampling.

\section{Benchmarks}
We selected eight notable methods in patent classification and extreme multi-label classification to benchmark against our datasets: FastXML \cite{fastxml}, PfastXML \cite{Jain2016ExtremeApplications}, Parabel \cite{Prabhu2018Parabel:Advertising}, CNN \cite{Kim2014ConvolutionalClassification}, XML-CNN \cite{Liu2017DeepClassification}, PatentBERT  \cite{Lee2019PatentBERT:Model}, LSAN \cite{Xiao2019Label-SpecificClassification}, and AttentionXML \cite{You2019AttentionXML:Classification}. FastXML, PfastXML, and Parabel are tree classifiers that efficiently employ the large label space in extreme multi-label classification. Particularly, FastXML and PfastXML create trees that partition the feature space, whereas Parabel does partitioning on the label space. 
On the other hand, deep neural networks have proven to be effective at document embedding. CNN, XML-CNN, PatentBERT, and LSAN learn to efficiently represent patent content for better classification. Both CNN and XML-CNN employ convolutional layers to encode textual information, whereas PatentBERT fine-tunes BERT, a model with stacked Transformer encoders, on patent datasets. Instead of a single representation, LSAN proposes a label-specific representation aligned with the purpose of patent classification. Combining the advantages of both approaches, AttentionXLM uses Bi-LSTM for patent embedding and utilizes a label tree for classification. We provide details of each method below.

\subsection{FastXML}
FastXML \cite{fastxml} constructs a tree that hierarchically partitions the feature space. Each node in the tree contains a learnable linear separator that assigns a sample to its left or right branch. The branching stops when the number of samples per leaf node remains below a threshold. For optimization, the authors suggest a new loss based on the difference between the sample labels and the top-$k$ most occurred labels at that leaf node. Intuitively, FastXML directly optimizes the ranking metrics $nDCG$ instead of traditional criteria such as Gini-index or clustering error.

\subsection{PfastXML}
Extreme multi-label data typically suffer from missing ground-truth labels, i.e., some round truth labels are not available due to the high cost of labeling. PfastXML \cite{Jain2016ExtremeApplications}, built upon FastXML, uses propensity to re-weight the loss function, resulting in an unbiased estimation of the true loss. The propensity score for class $l$ measures the probability of ground-truth label $l$ being observed in the data: 
$$p_l = P(y_l = 1 \mid y^*_l = 1)$$
where $y^*_l$ denotes complete (but unobtainable) labels and $y_l$ denotes observed (but possibly missing) labels of the class $l$. The authors used the sigmoid function to approximate propensity coefficients with respect to number of samples $N$ and other hyperparameters $A, B$.
$$p_{l} = \frac{1}{1 + \log(N-1)(B+1)^A e^{-A \log(N_l + B)}}$$

\subsection{Parabel}
Parabel \cite{Prabhu2018Parabel:Advertising} constructs a balanced tree that partitions the label space. A label is represented by the mean of its corresponding training sample embeddings. Each node in a tree contains two learnable linear classifiers which assign a sample to its left, right or both branches. The two classifiers reflect that a sample can belong to different label groups at the same time. There are 1-vs-all classifiers at leaf nodes, which explains the desire for a balanced tree. 

\subsection{CNN}
We implemented a simple baseline with CNN-based classification method \cite{Kim2014ConvolutionalClassification}, which uses convolutional layers to extract features from patent content. Specifically, we simultaneously used convolutional layers with different window sizes to extract features at different scales. Final document representation was obtained by concatenating convolutional outputs. A fully connected layer was used for classification.

\subsection{XML-CNN}
XML-CNN \cite{Liu2017DeepClassification} uses convolutional layers to extract features from patent content. Frequently, a convolutional layer is followed by a max-pooling layer, which keeps the largest values in the feature map. The authors argue its inefficiency for long document embedding and opt to use dynamic max pooling. A dynamic max-pooling layer retains $k$ largest values in the $k$ equal sub-map. Intuitively, it can capture more fine-grained features from different regions in a document. In addition, XML-CNN uses a fully-connected layer between the pooling and output layer as a second non-linear operation to enrich the document representation and reduce the number of parameters.

\subsection{PatentBERT}
In most recent years, BERT \cite{DCLT-NAACL-19} and its variants have been dominating architecture in natural language processing. PatentBERT \cite{Lee2019PatentBERT:Model} relies on the pre-trained BERT to fine-tune on patent datasets. At its core, BERT is a language model consisting of multiple stacked Transformers encoders enhanced with a bidirectional forward strategy to predict masked tokens. Each Transformer encoder includes a stacked multi-head attention layer, which enables the model to pay attention to different source tokens upon predicting a masked token with multiple possibilities. On the other hand, fine-tuning makes use of pre-trained knowledge acquired on a large corpus and adapts to the target domain. The authors of PatentBERT used pre-trained BERT weights and added a multi-label classifier on top for patent classification.

\subsection{LSAN}
LSAN \cite{Xiao2019Label-SpecificClassification} uses two strategies to obtain label-specific document representation. On one hand, it employs an attention mechanism to encode the document with respect to each label. On the other hand, it weights the universal document representation by its distance to label embeddings. Intuitively, the former focuses on patent content whereas the latter utilizes the semantic relation between labels and patents. LSAN learns to combine both representations to obtain label-specific representation, which is used for classification.

\begin{table*}[!h]
\caption{Performance comparison on two datasets. \textbf{Bold} and \underline{underlined} numbers the best and second best results.}
\label{tab:results}
\begin{tabular}{ccccccccc}\hline
\textbf{Data} & \textbf{Method} & \textbf{Micro F-1} & \textbf{nDCG@1} & \textbf{nDCG@3} & \textbf{nDCG@5} & \textbf{P@1} & \textbf{P@3} & \textbf{P@5} \\ \hline
\multirow{7}{*}{CinPatent-EN} 
    & FastXML & 37.41 & 65.80 & 64.53 & 67.54 & 65.80 & 38.99 & 27.63 \\
    & PfastXML & 37.74 & 65.29 & 64.47 & 67.58 & 65.30 & 39.08 & 27.72 \\
    & Parabel & 43.49 & \underline{75.02} & \underline{72.78} & \underline{75.53} & \underline{74.99} & \underline{43.53} & \underline{30.32} \\
    & CNN & 44.80 & 70.50 & 68.96 & 72.28 & 70.40 & 41.12 & 29.25 \\
    & XML-CNN & 44.79 & 70.08 & 68.47 & 71.60 & 70.08 & 40.79 & 28.89 \\
    & PatentBERT & \underline{53.38} & 74.09 & 70.78 & 73.28 & 74.09 & 41.65 & 28.83 \\
    & LSAN & 40.00 & 66.30 & 64.57 & 67.48 & 66.30 & 38.07 & 26.85 \\
    & AttentionXML & \textbf{58.86} & \textbf{78.07} & \textbf{76.13} & \textbf{79.12} & \textbf{78.07} & \textbf{45.63} & \textbf{32.01} \\ \hline

\multirow{7}{*}{CinPatent-JA} 
    & FastXML & 43.16 & 72.39 & 69.42 & 72.08 & 72.40 & 43.33 & 30.67 \\
    & PfastXML & 43.81 & 72.23 & 69.58 & 72.12 & 72.24 & 43.46 & 30.69 \\
    & Parabel & 46.88 & \underline{80.10} & \underline{76.39} & \underline{78.37} & \underline{80.10} & \underline{47.46} & \underline{32.88} \\
    & CNN & 46.09 & 71.79 & 68.68 & 71.14 & 71.79 & 42.13 & 29.74 \\
    & XML-CNN & 21.17 & 56.87 & 56.15 & 59.44 & 56.87 & 35.71 & 26.19 \\
    & PatentBERT & \underline{58.19} & 79.26 & 74.91 & 76.97 & 79.26 & 45.99 & 31.96 \\
    & LSAN & 55.09 & 74.71 & 70.20 & 72.35 & 74.71 & 42.76 & 29.87 \\
    & AttentionXML & \textbf{60.84} & \textbf{81.48} & \textbf{77.69} & \textbf{79.83} & \textbf{81.48} & \textbf{48.24} & \textbf{33.56} \\ \hline
\end{tabular}
\end{table*}

\subsection{AttentionXML}
AttentionXML \cite{You2019AttentionXML:Classification} is a hybrid of tree classifiers and deep neural networks. Specifically, it builds a probabilistic label tree as suggested in Parabel. There is a Bi-LSTM layer enhanced with multi-label attention that assigns a sample to one or multiple nodes at each level. In other words, a level in the tree is a multi-label subproblem. For better optimization, the authors attempted to keep the label tree wide and shallow with balanced $k$-means.

\section{Experiments and Results}

\subsection{Evaluation Metrics}
We followed common metrics of multi-label text classification for patent classification \cite{You2019AttentionXML:Classification,Xiao2019Label-SpecificClassification,Lee2019PatentBERT:Model}. The first metric is
Micro F-1 with the assumption that there is equivalent importance between precision and recall. The second set of metrics are ranking-based: Precision at top-$K$ (P@$K$) and Normalized Discounted Cumulative Gain at top-$K$ (NDCG@$K$).

\subsection{Implementation Details}
Most of the baseline implementations were taken from their official repositories, except for our implementations of CNN and PatentBERT. For each model, we initialized embeddings with recommended methods in the original papers from the concatenation of title+abstract+claim1+description. We explored different feature combinations in subsection \ref{subsection:results}. The concatenation length was limited to the first 5,000 tokens.

\paragraph{FastXML \& PfastXML}
We used TF-IDF features with the vocabulary size constrained within 50,000 tokens after ignoring English stopwords. Each token in the vocabulary appears at least 5 times and less than 80\% in the whole. We explicitly set 4 hyperparameters: number of trees for the ensemble as 5, maximum number of instances in a leaf node as 10, maximum number of leaf nodes with retained probabilities as 20, and the blending factor between left and right branch as 0.8. The propensity score was computed with A = 0.55 and B = 1.5. We trained both models in 20 epochs.

\paragraph{Parabel}
We used the same TF-IDF features as described in FastXML and PfastXML settings. We explicitly set 5 hyperparameters: number of trees for ensemble as 5, maximum number of instances in a leaf node as 100, maximum tree depth as 20, number of clusters in $k$-means as 2, and minimum number of instances in a cluster as 2. We train the model until specified tree depth was reached.

\paragraph{CNN}
We used word vectors from FastText \cite{bojanowski2017enriching} for the English dataset and randomized 300-dimensional vectors for the Japanese dataset. The model contains 4 parallel 1D CNN with kernel sizes [5, 4, 3, 2] and 128 filters. We trained CNN with the batch size of 64 in 29 epochs with a learning rate of $5e-4$ and the dropout rate of 0.2.

\paragraph{XML-CNN} 
We used GloVe 300-dimensional pre-trained vectors for the English dataset and randomized vectors with the size of 300 for the Japanese dataset. The model used rectified linear units as activation functions, 1-dimensional convolutional filters with the window sizes of 2, 4, 8; the number of feature maps for each filter was 128. We trained XML-CNN with the batch size of 64 in 31 epochs with a learning rate of $5e-4$ and the dropout rate of 0.5.
 
\paragraph{LSAN}
We used the sample input embedding method in XML-CNN for LSAN. Label embeddings were initialized randomly with 300-dimensional vectors. The model contains one bidirectional long short-term memory (Bi-LSTM) \cite{huang2015bidirectional} layer and label-word self-attention layers with the hidden size of 300. We trained LSAN with batch size 64 in 10 epochs with a learning rate of $1e-3$.

\paragraph{PatentBERT}
We used pre-trained weights and embeddings from HuggingFace Model Hub,\footnote{https://huggingface.co/models} particularly `bert-base-uncased' for English and `cl-tohoku/bert-base-japanese-char' for Japanese. Both models have 12 attention heads, 12 layers with the hidden size of each layer as 768 and maximum sequence length as 256. We used AdamW with epsilon 1e-8 and learning rate $5e-5$ for optimization. Both models were trained with batch size 32 in 30 epochs.

\paragraph{AttentionXML} 
We used the sample input embedding method in XML-CNN for LSAN.
The model contains a Bi-LSTM and linear layer with hidden size 256. We trained AttentionXML with the batch size of 32 in 19 epochs, the learning rate of $1e-3$. The dropout of the model was set as 0.5.

\subsection{Experimental Results} \label{subsection:results}
\paragraph{Performance comparison}
We provide results of eight baselines on our datasets in Table \ref{tab:results}. As we can observe that, AttentionXML \cite{You2019AttentionXML:Classification} is the best, in which it is consistently better than other strong methods. This is because AttentionXML combines a tree structure suggested in Parabel \cite{Prabhu2018Parabel:Advertising} with a deep neural network (Bi-LSTM) for learning hidden representation of patents. The model also takes into account label information by using attention. PatentBERT \cite{Lee2019PatentBERT:Model} and Parabel \cite{Prabhu2018Parabel:Advertising} are competitive in which they are the second best in several metrics. For PatentBERT, the model takes advantage of pre-trained BERT-like models for patent classification. The knowledge from pre-trained models (i.e., BERT) allows PatentBERT to effectively represent the contextual aspect of tokens, which are beneficial for fine-tuning on new samples in the patent domain. For Parabel, it utilizes the tree structure to build two classifiers that assign a sample into the branches of a node. These classifiers allow a sample can belong to different label groups at the same time. By using this structure, Parabel is the second best on almost all metrics. Results of other methods provide a good summary for the next studies on our datasets.

\paragraph{The ablation study}
We tested AttentionXML (due to its performance) with six combinations by using: (\textbf{1}) the title, (\textbf{2}) title+abstract, (\textbf{3}) title+abstract+claim1, (\textbf{4}) title+abstract+all claims, (\textbf{5}) title+abstract+all claims+description, and (\textbf{6}) title+abstract+claim1+description. We report six metrics due to the same values of $nDCG1$ and $P@1$.

\begin{table}[h!]
\caption{The ablation study with AttentionXML.}\label{tab:ablation}
\resizebox{\columnwidth}{!}{%
\begin{tabular}{lccccccc}
\hline
\textbf{Metrics} & \textbf{1} & \textbf{2} & \textbf{3} & \textbf{4} & \textbf{5} & \textbf{6} \\ \hline
\multicolumn{7}{c}{\textbf{CinPatent-EN}} \\ \hline
Micro F-1 & 19.94 & 44.19 & 46.43 & 47.78 & 54.88 & \textbf{58.86} \\
nDCG@3 & 38.64 & 59.07 & 62.15 & 63.61 & 72.27 & \textbf{76.13} \\
nDCG@5 & 41.18 & 62.04 & 64.87 & 66.47 & 75.13 & \textbf{79.12} \\
P@1 & 38.81 & 60.24 & 63.82 & 65.84 & 74.35 & \textbf{78.07} \\
P@3 & 23.24 & 35.21 & 37.11 & 37.74 & 43.15 & \textbf{45.63} \\
P@5 & 16.96 & 25.16 & 26.18 & 26.64 & 30.23 & \textbf{32.01} \\ \hline

\multicolumn{7}{c}{\textbf{CinPatent-JA}} \\ \hline
Micro F-1 & 48.17 & 48.27 & 50.39 & 54.98 & 55.28 & \textbf{60.84} \\
nDCG@3 & 62.75 & 62.84 & 65.81 & 70.76 & 71.46 & \textbf{77.69} \\
nDCG@5 & 64.98 & 65.11 & 68.16 & 73.29 & 73.80 & \textbf{79.83} \\
P@1 & 65.03 & 65.16 & 68.51 & 74.18 & 74.64 & \textbf{81.48} \\
P@3 & 38.30 & 38.28 & 40.32 & 43.54 & 44.09 & \textbf{48.24} \\
P@5 & 26.96 & 26.96 & 28.39 & 30.76 & 30.97 & \textbf{33.56} \\ \hline

\end{tabular}%
}
\vspace{-0.4cm}
\end{table}

In Table \ref{tab:ablation}, AttentionXML outputs the best results by using the sixth setting. This is because the combination of title, abstract, description, and claim1 provides good information to represent the content of patents. The results consistently increase when adding more information. For CinPatent-EN, the margins between (\textbf{1}) and (\textbf{2}) are significant. It shows that the abstract provides useful information. The setting (\textbf{5}) is much better than (\textbf{4}), showing that the description is also important. For CinPatent-JA, the gap between settings (\textbf{1}) and (\textbf{2}) is tiny. As shown in Table \ref{tb:data_stat_by_part}, there are 94.34\% of 54k Japanese patents which do not have abstract. The setting (\textbf{4}) is much better than (\textbf{3}), showing that using all claims is better than only using the first claim.

\begin{figure*}[!h]
    \centering
    \begin{subfigure}[b]{1.0\textwidth}
        \centering
        \includegraphics[width=0.98\textwidth]{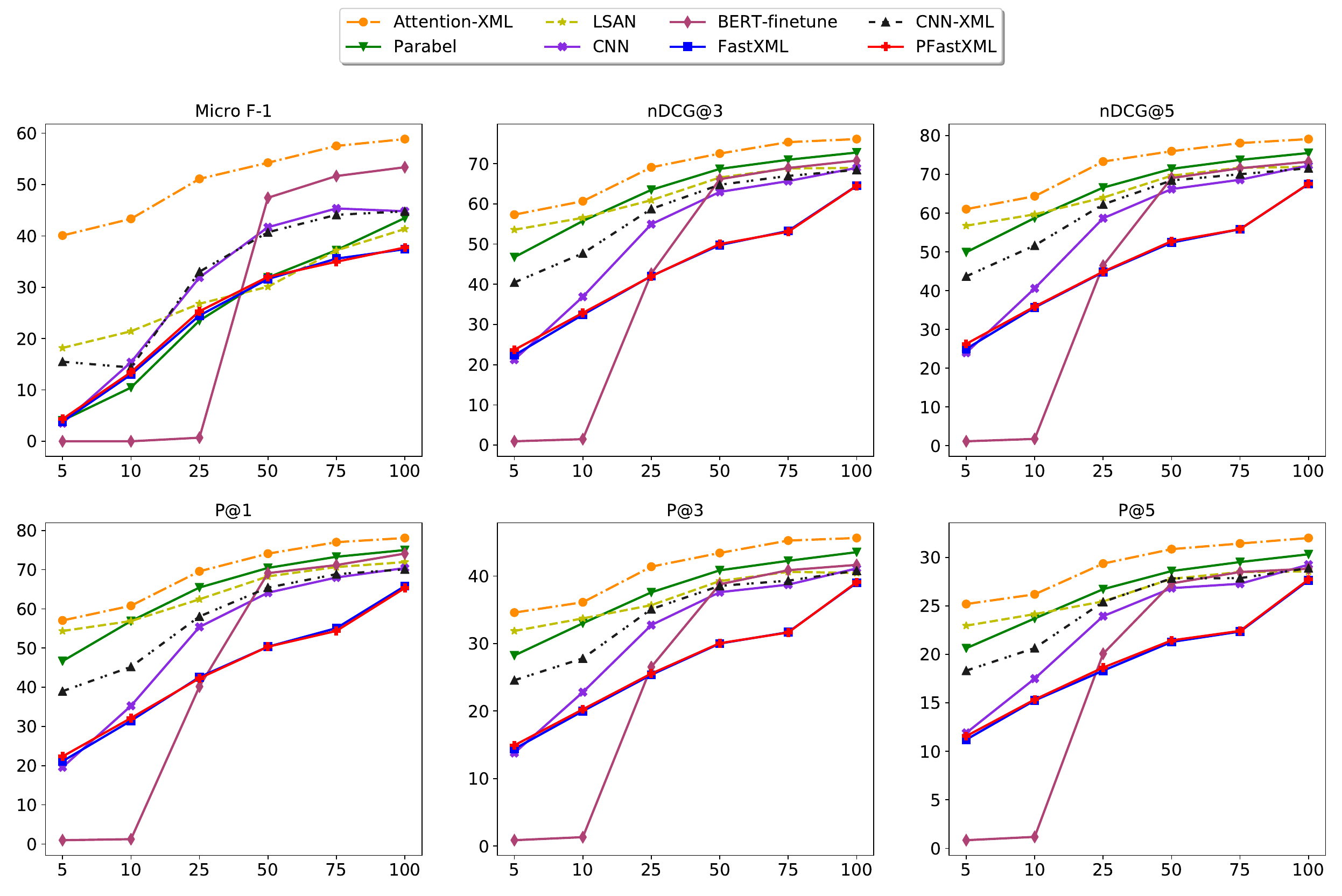}
    \end{subfigure}
    \caption{The performance of baselines with different data segmentation on CinPatent-EN.}
    \label{fig:en-line-graph}
    
    \begin{subfigure}[b]{1.0\textwidth}
        \centering
        \includegraphics[width=0.98\textwidth]{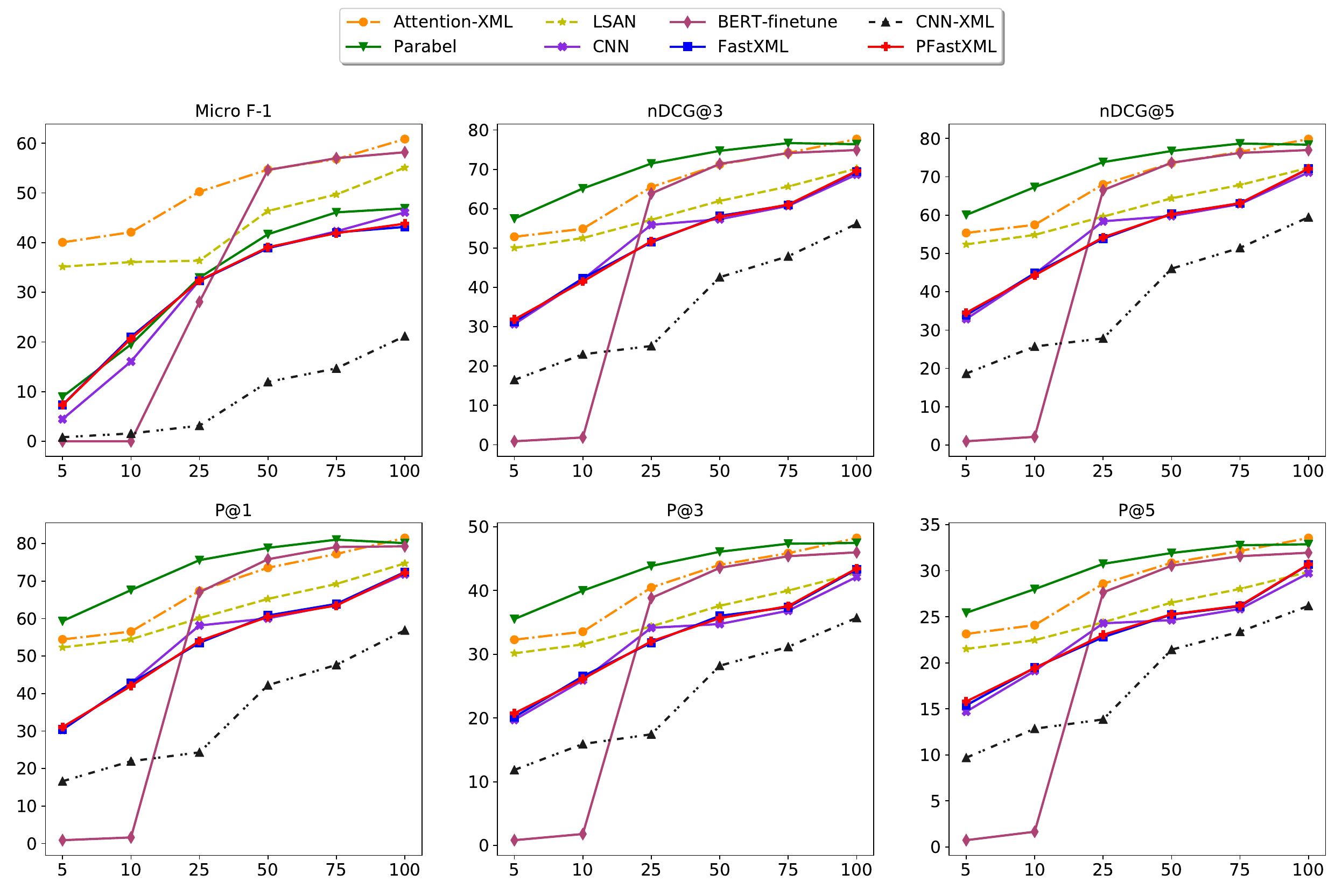}
    \end{subfigure}
    \caption{The performance of baselines with different data segmentation on CinPatent-JA.}
    \label{fig:jp-line-graph}
\end{figure*}

\paragraph{Performance with different data segmentation}
We investigated the behavior of baselines when training with different data segmentation. To do that, we trained all baselines on [5, 10, 25, 50, 75, 100] percentage of training samples. The trend in Figures \ref{fig:en-line-graph} and \ref{fig:jp-line-graph} are consistent with the results in Table \ref{tab:results}, in which AttentionXML using 100\% training samples is the best, followed by PatentBERT and Parabel. For CinPatent-EN, AttentionXNL is constantly better than other methods for all data segmentation. It again confirms its efficiency. The trend of Parabel is also consistent, which is the second best, except for Micro F-1. However, for CinPatent-JA, Parabel shows its efficiency for almost all metrics, except for Micro F-1. Parabel is better than AttentionXML over almost all data segmentation. It shows two interesting points. Firstly, Parabel may be appropriate for small data, compared to AttentionXML. Second, the performance of AttentionXML can be improved by adding more training samples.


\section{Conclusion}
This paper introduces two new datasets for patent classification in English and Japanese. The two datasets are validated by eight multi-label text classification methods. Experimental results show that AttentionXML is the best. The ablation study also indicates that the combination of title, abstract, the first claim, and description provides good information for AttentionXML. We encourage testing new methods for patent and multi-label text classification on our datasets.

\section*{Acknowledgement}

We would like to thank Phan Huy Tung for his help for data collection and pre-processing.


\bibliographystyle{lrec2022-bib}
\bibliography{references,patent}

\begin{thebibliography}{}

\bibitem[\protect\citename{Bojanowski \bgroup et al.\egroup
  }2017]{bojanowski2017enriching}
Bojanowski, P., Grave, E., Joulin, A., and Mikolov, T.
\newblock (2017).
\newblock Enriching word vectors with subword information.

\bibitem[\protect\citename{Degroote and Held}2018]{Degroote-Analysis-WPI-18}
Degroote, B. and Held, P.
\newblock (2018).
\newblock Analysis of the patent documentation coverage of the cpc in
  comparison with the ipc with a focus on asian documentation.
\newblock {\em World Patent Information 54: S78-S84}.

\bibitem[\protect\citename{Devlin \bgroup et al.\egroup }2018]{DCLT-NAACL-19}
Devlin, J., Chang, M.-W., Lee, K., and Toutanova, K.
\newblock (2018).
\newblock Bert: Pre-training of deep bidirectional transformers for language
  understanding.
\newblock In {\em Proceedings of the 2019 Conference of the North American
  Chapter of the Association for Computational Linguistics: Human Language
  Technologies, Volume 1 (Long and Short Papers)}, pages 4171--4186.

\bibitem[\protect\citename{Hepburn}2018]{Hepburn-ALTA-18}
Hepburn, J.
\newblock (2018).
\newblock Universal language model fine-tuning for patent classification.
\newblock In {\em Proceedings of the Australasian Language Technology
  Association Workshop 2018, pp. 93-96}.

\bibitem[\protect\citename{Hu \bgroup et al.\egroup
  }2018]{Hu-MPatent-Sustainability-18}
Hu, J., Li, S., Hu, J., and Yang, G.
\newblock (2018).
\newblock A hierarchical feature extraction model for multi-label mechanical
  patent classification.
\newblock {\em Sustainability 10, no. 1: 219}.

\bibitem[\protect\citename{Huang \bgroup et al.\egroup
  }2015]{huang2015bidirectional}
Huang, Z., Xu, W., and Yu, K.
\newblock (2015).
\newblock Bidirectional lstm-crf models for sequence tagging.

\bibitem[\protect\citename{Jain \bgroup et al.\egroup
  }2016]{Jain2016ExtremeApplications}
Jain, H., Prabhu, Y., and Varma, M.
\newblock (2016).
\newblock {Extreme multi-label loss functions for recommendation, tagging,
  ranking {\&} other missing label applications}.
\newblock In {\em Proceedings of the ACM SIGKDD International Conference on
  Knowledge Discovery and Data Mining}, volume 13-17-August-2016, pages
  935--944.

\bibitem[\protect\citename{Kim}2014]{Kim2014ConvolutionalClassification}
Kim, Y.
\newblock (2014).
\newblock {Convolutional neural networks for sentence classification}.
\newblock {\em 2014 Conference on Empirical Methods in Natural Language
  Processing, Proceedings of the Conference}, pages 1746--1751.

\bibitem[\protect\citename{Krestel \bgroup et al.\egroup
  }2021]{Krestel-Patent-analysis-WPI-21}
Krestel, R., Chikkamath, R., Hewel, C., and Risch, J.
\newblock (2021).
\newblock A survey on deep learning for patent analysis.
\newblock {\em World Patent Information 65: 102035}.

\bibitem[\protect\citename{Lee and Hsiang}2019]{Lee2019PatentBERT:Model}
Lee, J.-S. and Hsiang, J.
\newblock (2019).
\newblock {PatentBERT: Patent Classification with Fine-Tuning a pre-trained
  BERT Model}.
\newblock 1:1--6.

\bibitem[\protect\citename{Lee and Hsiang}2020]{Lee-UPTO-3M-20}
Lee, J.-S. and Hsiang, J.
\newblock (2020).
\newblock Patent classification by fine-tuning bert language model.
\newblock In {\em World Patent Information 61: 101965}.

\bibitem[\protect\citename{Li \bgroup et al.\egroup
  }2018]{Li-DeepPatent-Scientometrics-18}
Li, S., Hu, J., Cui, Y., and Hu, J.
\newblock (2018).
\newblock Deeppatent: patent classification with convolutional neural networks
  and word embedding.
\newblock {\em Scientometrics 117, no. 2: 721-744}.

\bibitem[\protect\citename{Lim and Kwon}2016]{Lim-IPC-ICADMA-16}
Lim, S. and Kwon, Y.
\newblock (2016).
\newblock Ipc multi-label classification based on the field functionality of
  patent documents.
\newblock In {\em International Conference on Advanced Data Mining and
  Applications, pp. 677-691. Springer, Cham}.

\bibitem[\protect\citename{Liu \bgroup et al.\egroup
  }2017]{Liu2017DeepClassification}
Liu, J., Chang, W.~C., Wu, Y., and Yang, Y.
\newblock (2017).
\newblock {Deep learning for extreme multi-label text classification}.
\newblock In {\em SIGIR 2017 - Proceedings of the 40th International ACM SIGIR
  Conference on Research and Development in Information Retrieval}, pages
  115--124. Association for Computing Machinery, Inc, 8.

\bibitem[\protect\citename{Mccann}2020]{Mccann2020FugashiPython}
Mccann, P.
\newblock (2020).
\newblock {fugashi, a Tool for Tokenizing Japanese in Python}.

\bibitem[\protect\citename{Prabhu and Varma}2014]{fastxml}
Prabhu, Y. and Varma, M.
\newblock (2014).
\newblock {FastXML: A fast, accurate and stable tree-classifier for extreme
  multi-label learning}.
\newblock {\em Proceedings of the ACM SIGKDD International Conference on
  Knowledge Discovery and Data Mining}, pages 263--272.

\bibitem[\protect\citename{Prabhu \bgroup et al.\egroup
  }2018]{Prabhu2018Parabel:Advertising}
Prabhu, Y., Kag, A., Harsola, S., Agrawal, R., and Varma, M.
\newblock (2018).
\newblock {Parabel: Partitioned label trees for extreme classification with
  application to dynamic search advertising}.
\newblock In {\em The Web Conference 2018 - Proceedings of the World Wide Web
  Conference, WWW 2018}, pages 993--1002. Association for Computing Machinery,
  Inc, 4.

\bibitem[\protect\citename{Pujari \bgroup et al.\egroup
  }2021]{Pujari-Mutil-task-Transformers-ECIR-21}
Pujari, S.~C., Friedrich, A., and Strötgen, J.
\newblock (2021).
\newblock A multi-task approach to neural multi-label hierarchical patent
  classification using transformers.
\newblock In {\em European Conference on Information Retrieval, pp. 513-528.
  Springer, Cham}.

\bibitem[\protect\citename{Tang \bgroup et al.\egroup
  }2020]{Tang-MLP-GCN-AAAI-20}
Tang, P., Jiang, M., Xia, B.~N., Pitera, J.~W., Welser, J., and Chawla, N.~V.
\newblock (2020).
\newblock Multi-label patent categorization with non-local attention-based
  graph convolutional network.
\newblock In {\em Proceedings of the AAAI Conference on Artificial
  Intelligence, vol. 34, no. 05, pp. 9024-9031}.

\bibitem[\protect\citename{Tran and Kavuluru}2017]{Tran-CPC-Codes-ICMIKE-17}
Tran, T. and Kavuluru, R.
\newblock (2017).
\newblock Supervised approaches to assign cooperative patent classification
  (cpc) codes to patents.
\newblock In {\em International Conference on Mining Intelligence and Knowledge
  Exploration, pp. 22-34. Springer, Cham}.

\bibitem[\protect\citename{Xiao \bgroup et al.\egroup
  }2019]{Xiao2019Label-SpecificClassification}
Xiao, L., Huang, X., Chen, B., and Jing, L.
\newblock (2019).
\newblock {Label-Specific Document Representation for Multi-Label Text
  Classification}.
\newblock Technical report.

\bibitem[\protect\citename{Yadrintsev \bgroup et al.\egroup
  }2018]{Yadrintsev-patent-search-JPC-18}
Yadrintsev, V., Bakarov, A., Suvorov, R., and Sochenkov, I.
\newblock (2018).
\newblock Fast and accurate patent classification in search engines.
\newblock {\em Journal of Physics: Conference Series, vol. 1117, no. 1, p.
  012004}.

\bibitem[\protect\citename{You \bgroup et al.\egroup
  }2019]{You2019AttentionXML:Classification}
You, R., Zhang, Z., Wang, Z., Dai, S., Mamitsuka, H., and Zhu, S.
\newblock (2019).
\newblock {AttentionXML: Label tree-based attention-aware deep model for
  high-performance extreme multi-label text classification}.
\newblock {\em Advances in Neural Information Processing Systems},
  32(NeurIPS):1--17.

\end{thebibliography}

\end{document}